\documentclass[runningheads]{llncs}
\usepackage[T1]{fontenc}
\usepackage{graphicx}
\usepackage{amsmath}
\usepackage{tabularx,booktabs}
\usepackage{subfigure}

\begin{document}
\title{SYNTHETIC ELECTRORETINOGRAM SIGNAL GENERATION USING CONDITIONAL GENERATIVE ADVERSARIAL NETWORK FOR ENHANCING CLASSIFICATION OF AUTISM SPECTRUM DISORDER}
%
%
\author{Mikhail Kulyabin\inst{1} \and
Paul A. Constable\inst{2} \and
Aleksei Zhdanov\inst{3} \and
Irene O. Lee\inst{4} \and
David H. Skuse\inst{4} \and
Dorothy A. Thompson\inst{5,6} \and
Andreas Maier\inst{1}}
\authorrunning{M. Kulyabin et al.}
%
\institute{Pattern Recognition Lab, University of Erlangen-Nuremberg, Germany \email{mikhail.kulyabin@fau.de}\\ \and
College of Nursing and Health Sciences, Caring Futures Institute, \\ Flinders University, Adelaide, Australia \and
Siemens Healthineers, Erlangen, Germany \and
Behavioural and Brain Sciences Unit, Population Policy and Practice Programme, UCL Great Ormond Street Institute of Child Health, \\ University College London, London, UK \and
The Tony Kriss Visual Electrophysiology Unit, \\ Clinical and Academic Department of Ophthalmology, \\ Great Ormond Street Hospital for Children NHS Trust, London, UK \and
UCL Great Ormond Street Institute of Child Health, \\ University College London, London, UK
}

\titlerunning{SYNTHETIC ELECTRORETINOGRAM SIGNAL GENERATION}

\maketitle              
\begin{abstract}
The electroretinogram (ERG) is a clinical test that records the retina's electrical response to light. The ERG is a promising way to study different neurodevelopmental and neurodegenerative disorders, including autism spectrum disorder (ASD) - a neurodevelopmental condition that impacts language, communication, and reciprocal social interactions. However, in heterogeneous populations, such as ASD, where the ability to collect large datasets is limited, the application of artificial intelligence (AI) is complicated. Synthetic ERG signals generated from real ERG recordings carry similar information as natural ERGs and, therefore, could be used as an extension for natural data to increase datasets so that AI applications can be fully utilized. As proof of principle, this study presents a Generative Adversarial Network capable of generating synthetic ERG signals of children with ASD and typically developing control individuals. We applied a Time Series Transformer and Visual Transformer with Continuous Wavelet Transform to enhance classification results on the extended synthetic signals dataset. This approach may support classification models in related psychiatric conditions where the ERG may help classify disorders.
\end{abstract}
\section{Introduction}
The electroretinogram (ERG) is the waveform recorded from the eye under dark- or light-adapted conditions in response to a brief flash of light. The stimulus parameters for the recording of the ERG are described by the International Society for Clinical Electrophysiology of Vision (ISCEV) standard, which details flash strength, duration, amplification, patient preparation, and reporting of the clinical ERG \cite{Robson_Frishman_Grigg_Hamilton_Jeffrey_Kondo_Li_McCulloch_2022}. Typically, the clinical ERG is used to support the diagnosis of inherited or acquired retinal disease using time-domain parameters based on the amplitude and timing of the waveform peaks \cite{Robson_Nilsson_Li_Jalali_Fulton_Tormene_Holder_Brodie_2018}. However, in conditions where subtle changes in the ERG may not be apparent through analysis of time-domain parameters, such as in heterogeneous conditions, that may co-occur, such as autism spectrum disorder (ASD) \cite{waterhouse2022heterogeneity} and attention deficit hyperactivity disorder (ADHD) \cite{antshel2019autism} then alternative approaches may be required using signal analysis of the ERG to identify features for better classification \cite{Constable_Marmolejo_Ramos_Gauthier_Lee_Skuse_Thompson_2022,manjur2024detecting}. One way of addressing the difficulty of heterogeneity is by using large datasets and artificial intelligence (AI) to improve classification models. To support this aim, the generation of synthetic ERGs may help to advance this field for the identification and classification of neurodevelopmental and neurodegenerative disorders where the ERG is anomalous \cite{Constable_Lim_Thompson_2023}.  

The amplitude of the ERG waveform is affected by various patient factors, including age \cite{Neveu_Dangour_Allen_Robson_Bird_Uauy_Holder_2011}, sex \cite{Brûlé_Lavoie_Casanova_Lachapelle_Hébert_2007}, iris color \cite{Al_Abdlseaed_McTaggart_Ramage_Hamilton_McCulloch_2010}, pupil diameter \cite{Davis_Kraszewska_Manning_2017}, dark adaptation interval \cite{Bach_Meroni_Heinrich_2020}, and in the case of skin electrodes, their position below the eye \cite{Hobby_Kozareva_Yonova-Doing_Hossain_Katta_Huntjens_Hammond_Binns_Mahroo_2018}. In addition to these subject factors, the choice of electrode, fiber, gold foil, contact lens, or skin will also affect the amplitude of the recorded signal \cite{Chen_Greenstein_Brodie_2022}. Consequently, the ISCEV standard recommends that clinical sites establish reference ranges that reflect the local population and recording parameters \cite{Robson_Frishman_Grigg_Hamilton_Jeffrey_Kondo_Li_McCulloch_2022}. Decomposing the signal may mitigate against some of these effects, such as age and iris color, by providing features that are less dependent on the patient's age, ocular pigmentation, and the testing parameters. Benefits of signal analysis have been demonstrated in ECG recordings \cite{yi2000predictive} but is yet to be routinely used in ophthalmology to analyze the ERG in detail and enhance the potential of the ERG as an early marker for retinal or neurological diseases \cite{Mahroo_2023}. 

Through demonstration of harnessing the utility of synthetic waveform generation to the ERG, this technique could support fields of ophthalmic research. These may include, but are not limited to, clinical trials to ensure appropriate balancing between case and control groups based on age, sex, electrode type, and iris color. In addition, the generation of synthetic ERG waveforms based on signals from natural data recorded in different populations would help with ensuring comparison reference waveforms are available to sites regardless of their local patient population and contribute to the expansion of the clinical utility of the ERG \cite{kulyabin2024generating}. Furthermore, the ERG is only one test that is often used in conjunction with other tests of the visual pathways that may also be amenable to synthetic waveform generation, such as the multifocal and pattern ERG \cite{Robson_Nilsson_Li_Jalali_Fulton_Tormene_Holder_Brodie_2018}. Thus, this paper intends to introduce this field to ophthalmology, using the ERG and autism as models from which future studies may build.

The ASD and control populations used in this study have been previously described in detail \cite{Constable_Marmolejo_Ramos_Gauthier_Lee_Skuse_Thompson_2022}. The ASD participants met DSM-5 classification based on clinical assessments at two sites. Here, we present a novel approach for generating and classifying synthetic ERG signals. The key contributions of this work encompass the following: (1) The introduction of a Generative Adversarial Network (GAN) to produce synthetic ERG waveforms corresponding to both ASD and control classes; (2) The application of a Visual Transformer (ViT) and a Time Series Transformer (TST), which are trained on the ERG signals in the time-frequency and time domains respectfully to classify the signals as either ASD or control. The synthetic data generated exhibited characteristics similar to natural ERG waves, and improved the performance of the classification model. Unlike other biological signals, such as the ECG, this approach has not been applied thoroughly to the ERG signal. Thus, this new approach to handling limited-sized datasets that utilize the ERG for disease classification or early detection in psychiatric disorders \cite{schwitzer2022using} will provide a new methodological approach to this research field.

\section{Related Work}
The field of medical signals generation with AI has received significant attention in recent years, prompting numerous advancements and this being the first work to propose a synthetic ERG  generation method using AI for ASD classification. Despite the previous significant body of work in this field, such as synthesizing electrocardiogram (ECG) and electroencephalogram (EEG) signals in one-dimensional space, no other studies have sought to utilize these methods for classifying neurodevelopmental disorders using the ERG signal. Previous works in neurology include those of Hartmann et al. \cite{Hartmann2018EEGGANGA}, who utilized EEG-GAN as a framework to generate EEG brain signals. They reported that the modification of the  Wasserstein GAN stabilized training and were then able to investigate a range of architectural choices critical for time series generation. In cardiology, Golany et al. \cite{golany2020improving} applied a DCGAN architecture to generate synthetic ECG signals independently to five different heartbeat classes. The authors showed that adding generated signals to the LSTM classifier improved accuracy significantly. Wang et al. \cite{8771369} augmented an imbalanced ECG dataset using signals generated by a 1D auxiliary classifier generative adversarial network (AC-GAN). In the same field, 
Zhu et al. \cite{Zhu_Ye_Fu_Liu_Shen_2019} employed a Long Short-Term Memory (LSTM) layer in GAN to generate synthetic ECG signals.
Thambawita et al. \cite{Thambawita_Isaksen_Hicks_Ghouse_Ahlberg_Linneberg_Grarup_Ellervik_Olesen_Hansen_et} developed and compared two methods, named WaveGAN and Pulse2Pulse, based on GAN architecture. According to the authors, conditional GAN (CGAN) improved ECG signal generation by paying more attention to minor signal features' importance. 

Most recently, we have used GAN to increase the dataset sample size in an under-represented class based on sex in a control population using standard light-adapted ERGs as a precursor to applying these methods to a classification framework that we present here \cite{kulyabin2024generating}.

In this study, we incorporated the concept of integrating a Bidirectional Long Short-Term Memory (BLSTM) layer into GAN architecture. We used a conditional mechanism to enhance the generation process as an additional improvement.

The idea of using machine learning (ML) algorithms to classify ASD ERG signals has been applied in the papers of Manjur et al. \cite{manjur2022detecting,manjur2024detecting}. Time-domain features could detect ASD with a maximum of 65\% accuracy. The classification accuracy of the authors' best ML model using time-domain and spectral features was 86\%, with 98\% sensitivity.

\section{Dataset}

The original dataset \cite{Constable2022data} is shown in Table \ref{tab:dataset}. It includes nine types of LA-ERG waveform recordings at different flash strengths, including 1.204, 1.114, 0.949, 0.799, 0.602, 0.398, 0.114, -0.119, -0.367 (\(log\; cd.s.m^{-2}\)) from 30 ASD and 20 typically developing control individuals.
Data was collected at two sites based in London (UK) and Adelaide (Australia) as described previously in detail in previous works \cite{Constable_Marmolejo_Ramos_Gauthier_Lee_Skuse_Thompson_2022,lee2022electroretinogram,constable2020light}.

\begin{table}[h]
\centering
\caption{Dataset distribution}
\label{tab:dataset}
\begin{tabular}{cc|cc}
\toprule
\multicolumn{2}{c|}{Flash Strength}  & \multicolumn{2}{c}{Signal Number}     \\
\midrule
Td.s & \(log\; cd.s.m^{-2}\)           & ASD          & Control          \\
\midrule
446 & 1.204          & 58           & 59               \\
356 & 1.114          & 60           & 51               \\
251 & 0.949          & 56           & 51               \\
178 & 0.799          & 56           & 57               \\
113 & 0.602          & 58           & 56               \\
70 & 0.398          & 60           & 55               \\
35 & 0.114          & 53           & 50               \\
21 & -0.119         & 56           & 50               \\
12 & -0.367         & 52           & 53               \\
\midrule
Total &           & 509          & 482              \\
\bottomrule
\end{tabular}
\end{table}

Full-field LA-ERG recordings were performed on each eye (right always first), and followed the guidelines of the ISCEV ERG standard \cite{Robson_Frishman_Grigg_Hamilton_Jeffrey_Kondo_Li_McCulloch_2022}. A series of white flashes at each strength were presented to the eye at 2 Hz on a 40 (\(cd.m^{-2}\)) white background. Recordings were performed with the RETeval (LKC Technologies, Gaithersburg, MD, USA) with a custom nine-step randomized Troland-based protocol with skin electrodes placed 2-3 $mm$ below the lower eyelid. ERG waveforms were averaged from 30-60 traces per eye to generate the reported averaged waveform signal that was used in the analysis. Waveforms with artifacts such as blinks were automatically rejected if they fell within the upper or lower quartile of the overall average. Two replicate recordings were typically made in each eye and were included in the dataset for analysis. All signals have a length of 235.

\section{Method}
\begin{figure}[htb]
\begin{minipage}[b]{1.0\linewidth}
  \centering
  \centerline{\includegraphics[width=10.0cm]{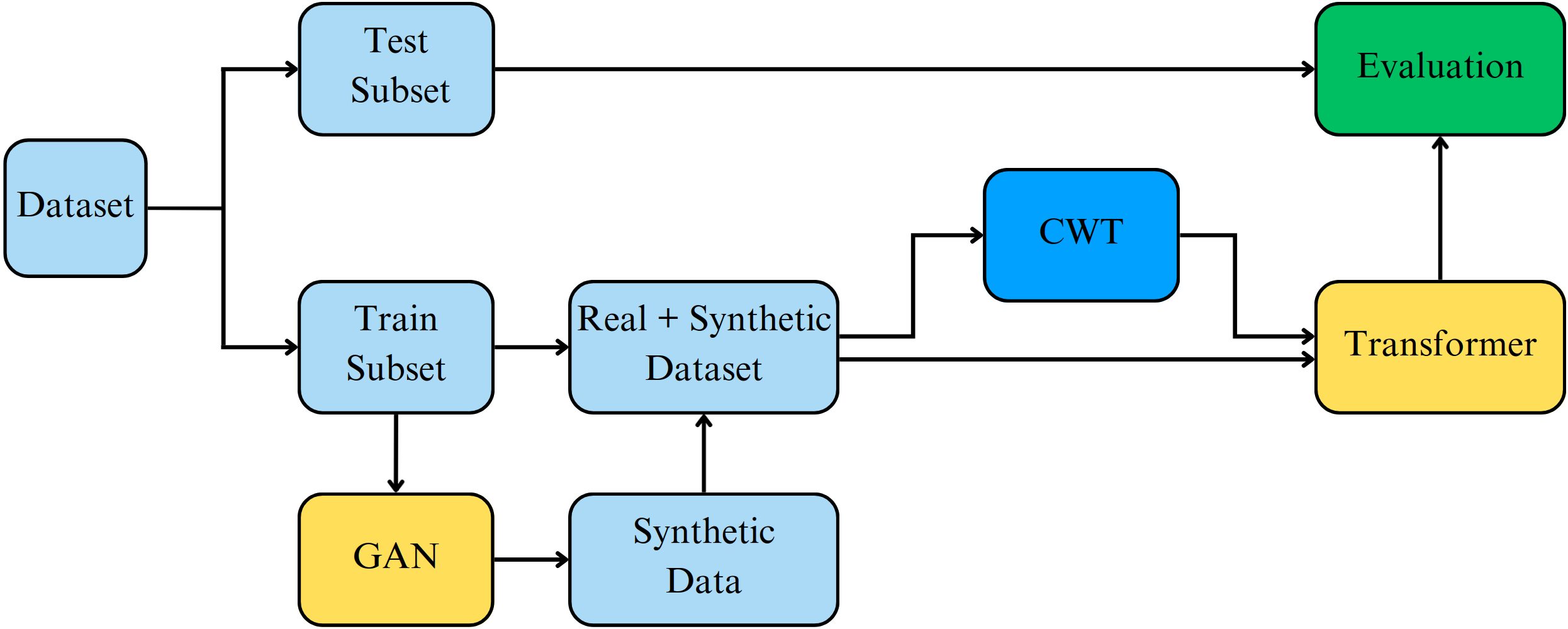}}
\end{minipage}
\caption{Proposed method overview. The dataset was split into test and training subsets so that the generative model was not trained on data correlated to the test for a fair evaluation. Synthetic signals generated by the generative model were added to real data. Classification models were then trained on merged and unmerged datasets in the time-frequency (ViT) and time (TST) domains.}
\label{fig:method}
\end{figure}

In this section, we present the proposed method for the generation of synthetic ERG signals and their further classification, shown in Fig. \ref{fig:method}. Initially, 25 \% of the data was partitioned as a test dataset and stored untouched for unbiased evaluation outcomes. The remaining subset was utilized to train the conditional GAN for synthesizing ERG signals. These generated signals were then combined with real or 'natural' signals to form an extended dataset for training the classification Transformer models. For the training of ViT, a CWT transformation was applied, generating a wavelet representation for each signal. For the TST training, signals were used in their original time-series representation. The evaluation process used a five-fold cross-validation, and the performance metrics were averaged over the test subsets.

\subsection{Conditional GAN}
\label{ssec:gan}

The framework for synthetic waveform generation used CGAN \cite{mirza2014conditional} to obtain the synthetic signals from the natural ERG waveforms. The CGAN architecture comprised two sub-networks: Generator (G) and Discriminator (D). The goal of the Generator was to learn the transformation between the latent distribution \(p_{z}\) and the real-world data distribution \(p_{d}\). The Discriminator learned to distinguish real signals from synthesized ones. The Generator and the Discriminator were provided with auxiliary class information \(y\) as an additional input layer. The complete CGAN architecture was trained in a min-max optimization game as in the standard GAN loss function: the Discriminator tried to maximize the score for real signals \(D(x|y)\) and minimize the score for generated \(D(G(z|y))\); and vice versa, the Generator tried to minimize \(log(1-D(G(z|y)))\), such that the minimizing of the term was only possible if the Generator synthesized realistic synthetic signals. The only difference with standard GAN is that the conditional probability was used for both the Generator and the Discriminator instead of the regular one. The complete optimization process is represented below \cite{mirza2014conditional}: 

\begin{equation}   
\underset{G}{min}\underset{D}{max}V(D,G) = E_{x{\sim}p_{d}(x)}[logD(x|y)] + E_{z{\sim}p_{z}(z)}[log(1-D(G(z|y)))]
\label{eq_gan}
\end{equation}

\begin{figure}[htb]

\begin{minipage}[b]{1.0\linewidth}
  \centering
  \centerline{\includegraphics[width=8cm]{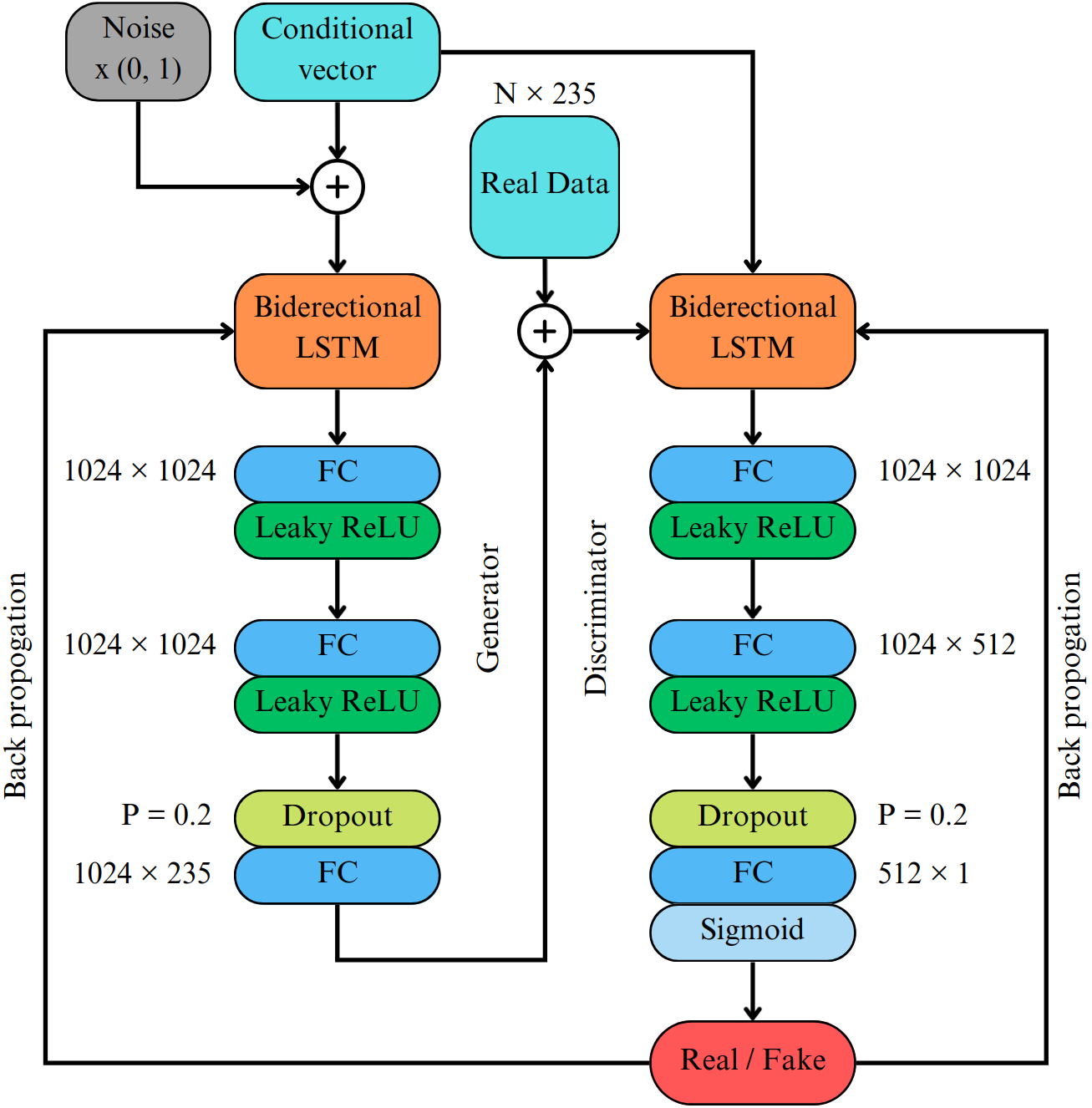}}
\end{minipage}

\caption{Conditional generative adversarial network (CGAN) structure.}
\label{fig:cgan}
\end{figure}

Recurrent Neural Networks (RNNs) have gained widespread adoption in diverse applications, including time series data processing, speech recognition, and image generation \cite{sherstinsky2020fundamentals}. While RNNs exhibit proficiency in handling short-term dependencies, they have limitations in effectively addressing long-term dependencies. To overcome these limitations, LSTM networks were introduced as an extension of RNNs \cite{hochreiter1997long}. LSTM incorporates a memory cell architecture that retains prior contextual information, thus accommodating issues such as gradient expansion or disappearance during training.

BLSTM is a neural network technique that facilitates bidirectional sequence information flow, encompassing both backward and forward directions \cite{graves2005framewise}. In contrast to the conventional LSTM architecture, where the input is unidirectional, either flowing backward or forward, BLSTM allows for concurrent information propagation in both directions. This bidirectional flow ensures the preservation and utilization of future and past contextual cues within the network, distinguishing BLSTM from its unidirectional LSTM counterpart. 

As shown in Fig. \ref{fig:cgan}, the Generator consists of a BLSTM layer of size $235 \times 512$, two blocks of fully connected (FC) layers of size $1024 \times 1024$, and a Leaky ReLU activation function. Then, dropout is followed by another FC layer of size $1024 \times 235$. The Discriminator has the same structure except for the sizes of FC layers ($1024 \times 1024$, $1024 \times 512$, and $512 \times 1$) and an additional Sigmoid function to scale the output in the range of 0 to 1.

\begin{figure*}[htp]
  \centering
  \subfigure[0.114 flash strength.]{\includegraphics[width=10cm]{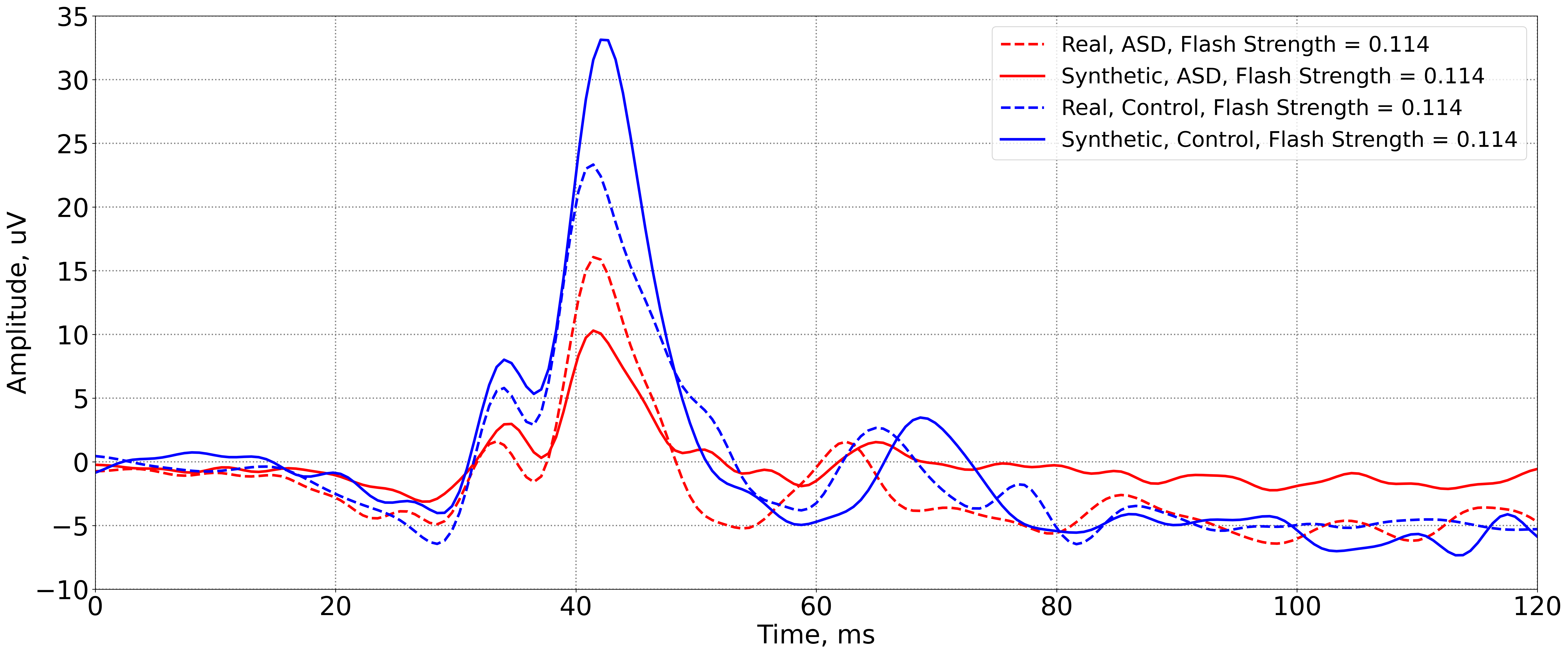}}
  \subfigure[- 0.119 flash strength.]{\includegraphics[width=10cm]{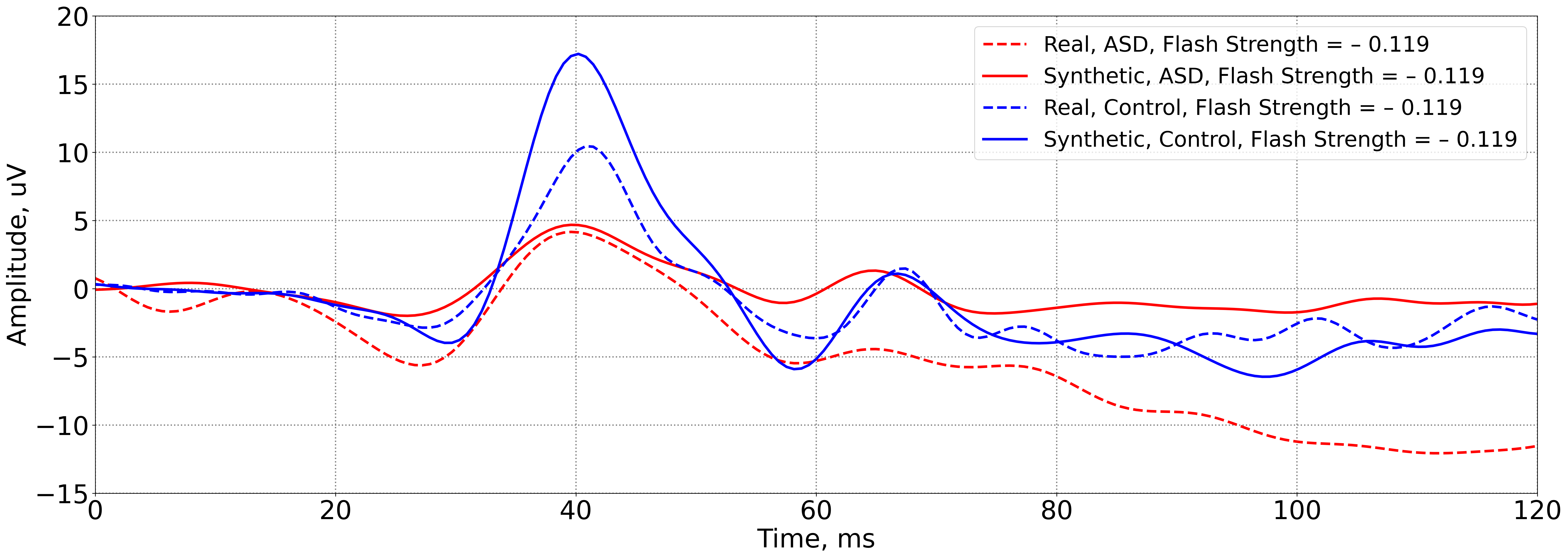}}
  \subfigure[- 0.367 flash strength.]{\includegraphics[width=10cm]{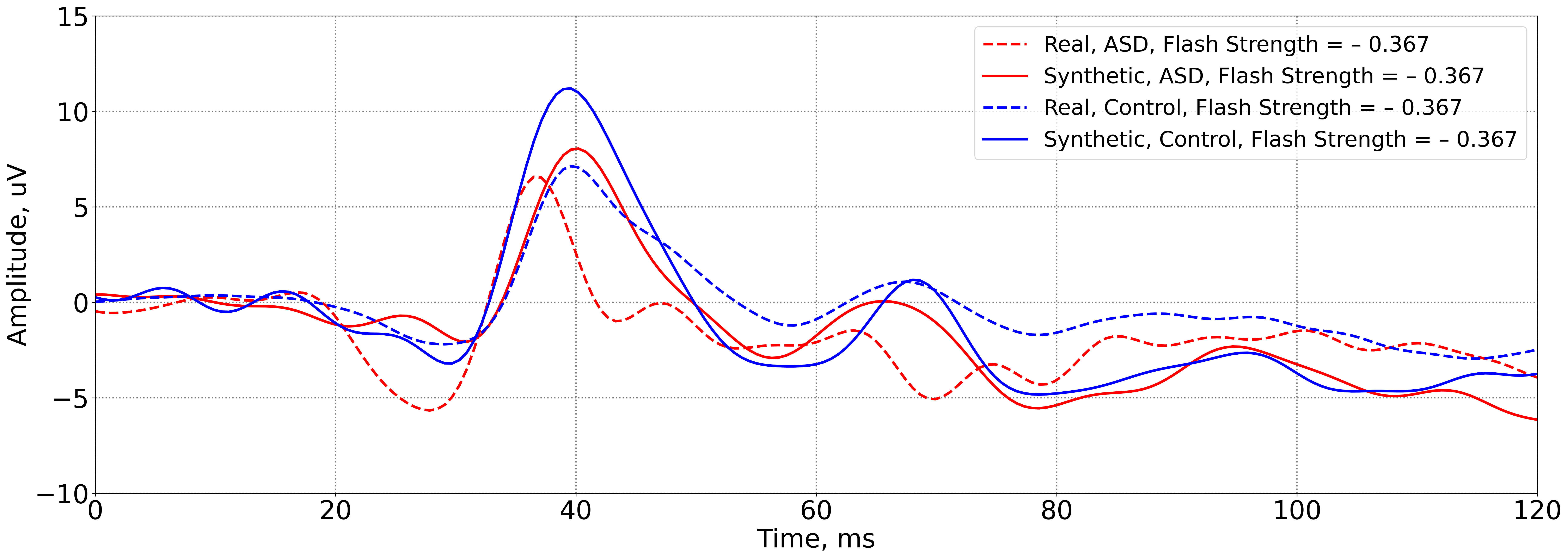}}
  \subfigure[1.204 flash strength.]{\includegraphics[width=10cm]{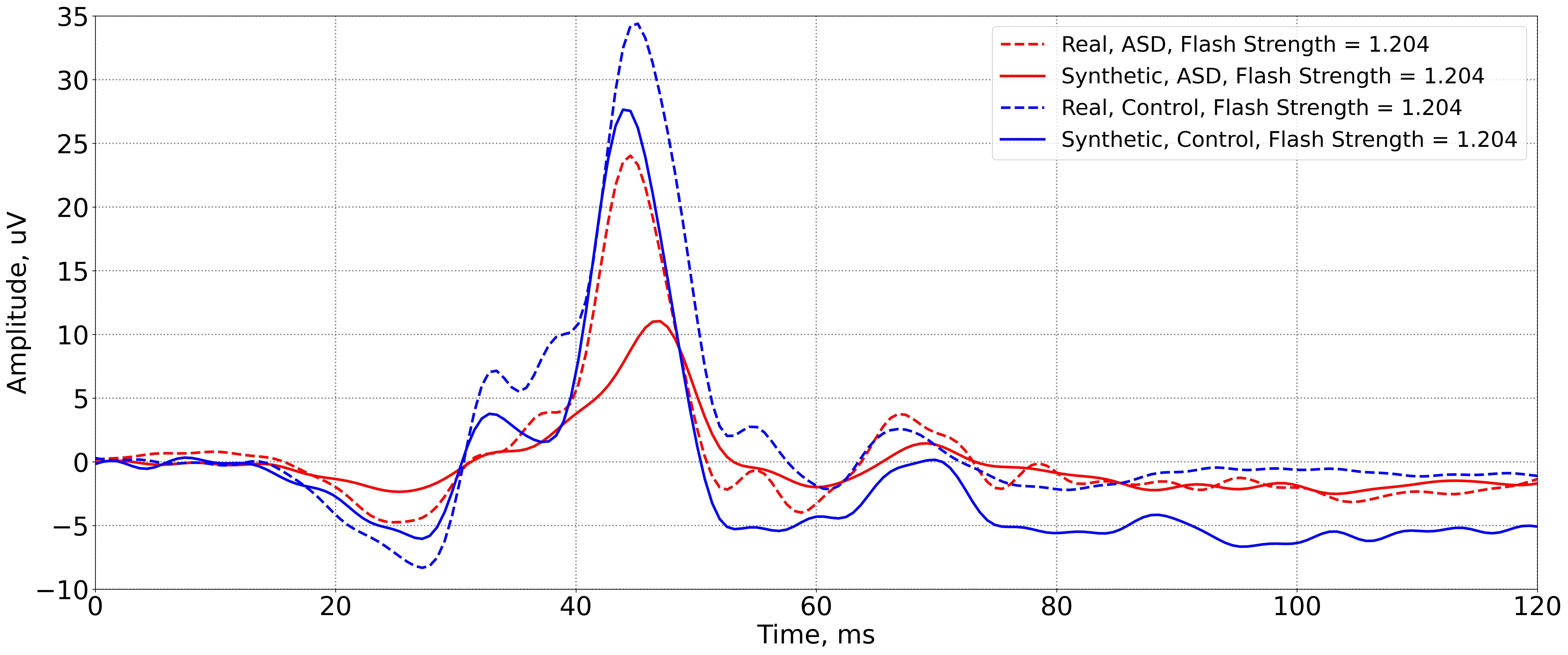}}
\caption{Examples of the real and synthetically generated by CGAN ERG signals for (a) 0.114, (b) - 0.119, (c) - 0.367, (d) 1.204 flash strengths \((log\: cd.s.m^{-2})\). Solid color lines represent synthetic ERG signals, dashed lines are real signals. Blue color corresponds to control, red color to ASD signals.}
\label{fig:generatedsignals}
\end{figure*}

Using the proposed model, hundreds of signals were synthesized for each class of ASD or control. Examples of the generated synthetic and real signals are shown in Fig. \ref{fig:generatedsignals}. The solid color lines represent synthetic ERG signals, and the dashed black lines are the original natural signals.

\subsection{Continuous Wavelet Transform}

Continuous Wavelet Transform (CWT) is an instrument that provides an overcomplete representation of a signal by letting the translation and scale parameter of the wavelets vary continuously. CWT of the function \(x(t)\) at \(a\) scale \((a>0)\in R^{+\ast}\) and translational value \(b\in R\) is expressed by the following integral (\ref{eq_cwt}), where \(\psi(t)\) is the continuous function called the mother wavelet, and the overline represents the operation of the complex conjugate \cite{grossmann_cwt}. The primary objective of the mother wavelet is to serve as the foundational function for generating daughter wavelets, which are simply the translated and scaled versions of the mother wavelet. The output of the CWT consists of a two-dimensional time-scale representation of the signal.

\begin{equation}   
X_{w}(a,b) = \frac{1}{|a|^{1/2}} \int^\infty_{-\infty} x(t) \overline{\psi} \left(\frac{t-b}{a}\right) dt
\label{eq_cwt}
\end{equation}

Using the method \cite{s23135813}, we determined the three most optimal mother functions for our dataset: Ricker, Gaussian, and Morlet. To increase the efficiency \cite{arias2020multi}, the input wavelet image for further classification consisted of three wavelets as three channels \cite{s23218727}.

\subsection{Visual Transformer classification model}
Transformers have emerged as a preferred model for image classification tasks, primarily attributed to their computational efficiency and scalability \cite{NIPS2017_3f5ee243}. 
In this study, we applied ViT to the wavelet scalograms using real and synthetic datasets. ViT demonstrated superior classification performance compared to training solely on the smaller natural waveform dataset.

\begin{figure}[h]
\begin{minipage}[b]{1.0\linewidth}
  \centering
  \centerline{\includegraphics[width=12cm]{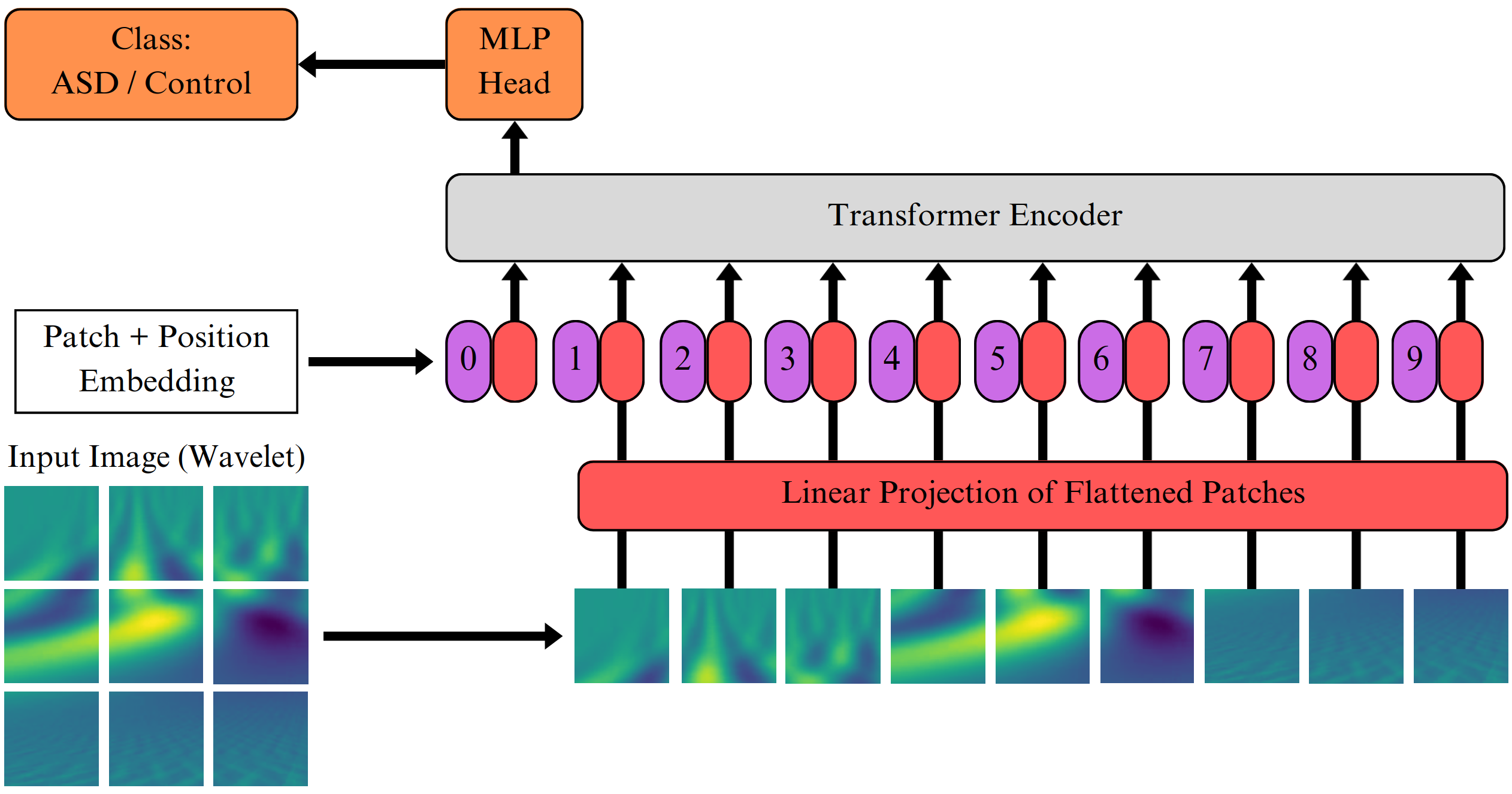}}
\end{minipage}
\caption{Overview of the Visual Transformer (ViT) architecture.}
\label{fig:vit}
\end{figure}

Fig. \ref{fig:vit} illustrates the ViT architecture proposed by Dosovitskiy et al. in 2020 \cite{dosovitskiy2020image}. Initially, the model processes a 2D input image (wavelet in our case) by transforming it into sequences of flattened 2D patches. These patches undergo a trainable linear projection to map them into a constant latent vector size. Before processing the patches through the encoder, a learnable embedding is added at the beginning of the sequence. For the classification task, the image representation is then passed through a classification head for fine-tuning. Position embeddings were then added to retain positional information, and the sequence of embedding vectors served as inputs to the Transformer encoder, which consisted of alternating layers of multiheaded self-attention and multilayer perceptron blocks \cite{NIPS2017_3f5ee243}.

The input sequence can be formed from feature maps of a convolution neural network (CNN) instead of raw image patches by leveraging a convolutional backbone \cite{dosovitskiy2020image}. In this case, the patch embedding projection was applied to patches extracted from a CNN feature map. This hybrid approach (ResNet - ViT in our case) enhanced the model's capability of handling smaller-sized images and improved parameter efficiency.

\subsection{Time Series Transformer classification model}

TST is a neural network architecture designed to process and analyze time series data. It is based on the Transformer architecture introduced by Vaswani et al. for natural language processing tasks \cite{NIPS2017_3f5ee243}, but without the decoder part of the architecture \cite{10.1145/3447548.3467401}.
TST adapts the Transformer architecture to handle sequences of temporal data points effectively. As with the original Transformer model, TST uses self-attention mechanisms to weigh the importance of different elements within the input time series sequence. This attention mechanism allows the model to learn dependencies between different time steps and capture long-range dependencies. Positional encodings are crucial to TST and are added to the input data to provide further information about the sequence's order of the time steps. The model also employed a multi-head attention mechanism to simultaneously capture different relationships and representations within the time series data. 

\subsection{Training}

CGAN was trained with a batch size of 15 on 10000 epochs. The dropout was set to 0.2. The Adam optimizer was used with a learning rate of 0.0002 for the Generator and Discriminator, with the BCELoss used as the criterion function. The training time of the CGAN model was two hours using an AMD Ryzen 95900HX×16 processor with an NVIDIA GeForce RTX 3070 graphics card. The model had $10^6$ trainable parameters with a complexity of 0.026 GFLOPS. 

For the classification, two ResNet - ViT hybrid image models were used, which differed in the number of parameters used and have shown effectiveness in ERG wavelet classification previously \cite{s23218727}. The models were ViT S (ViT\_small\_r26\_s32\_224) and ViT T (ViT\_tiny\_r\_s16\_p8\_224), available at the HuggingFace 'transformers' repository \cite{wolf2020transformers}. Both models were pretrained on ImageNet-21k and fine-tuned with ImageNet-1k with additional augmentation and regularization with a resolution of 224×224 pixels. An SGD optimization with a 0.001 initial learning rate was used. ViT S and ViT T have $36.4 \times 10^6$ and $10.4 \times 10^6$ trainable parameters with complexities of 3.5 and 0.4 GFLOPS, respectively. For TST, a cross-entropy loss function with class weights adapted to address the class imbalance and an Adam optimizer with an initial learning rate of 0.0001 was used. Each model was then trained until convergence using the early stopping criteria on the validation loss with a batch size of 32. 

The models were evaluated using a five-fold cross-validation. Performance metrics were Balanced Accuracy (BA), Precision (P), Recall (R), F1-score, and AUC. The performance outcomes were averaged across the five-folds and are presented in Table \ref{tab:results}. TST was trained for each strength independently, as well as on all strengths simultaneously. ViT was trained only on the entire signal dataset (all flash strengths together). The input for the ViT model was the wavelet scalograms obtained using CWT, and the input for TST was the ERG signals in the time-series representation.

\begin{table}[h]
\centering
\caption{Evaluation metrics of DL models}
\label{tab:results}
\begin{tabular}{lc | ccccc | ccccc}
\toprule
        &      & \multicolumn{5}{c}{Original} & \multicolumn{5}{c}{Original + Synthetic}   \\
\midrule
Network & Strength & BA & P & R & F1 & AUC & BA & P & R & F1 & AUC  \\  
\midrule
TST    &   1.204   & 0.710  & 0.710  & 0.736 & 0.717 & 0.739   & 0.842  & 0.933  & 0.736 & 0.823 & 0.897  \\
       &   1.114   & \textbf{0.805}  & \textbf{0.789}  & 0.833 & \textbf{0.810} & 0.768   & \textbf{0.891}  & \textbf{0.900}  & \textbf{0.947} & \textbf{0.923} & 0.916  \\
       &   0.949   & 0.759  & 0.727  & \textbf{0.888} & \textbf{0.799} & \textbf{0.845}   & \textbf{0.894}  & 0.857  & \textbf{0.947} & \textbf{0.914} & \textbf{0.969}  \\
       &   0.799   & 0.722  & 0.700  & 0.777 & 0.736 & 0.753   & 0.815  & 0.772  & 0.894 & 0.829 & 0.850  \\
       &   0.602   & 0.722  & 0.750  & 0.666 & 0.705 & 0.842   & 0.842  & 0.842  & 0.842 & 0.842 & 0.950  \\
       &   0.398   & 0.712  & 0.710  & 0.737 & 0.731 & 0.725   & 0.815  & 0.815  & 0.731 & 0.774 & 0.955  \\
       &   0.114   & 0.749  & 0.723  & 0.731 & 0.749 & 0.744   & 0.763  & 0.812  & 0.784 & 0.742 & 0.817  \\
       &  -0.119   & 0.750  & 0.736  & 0.777 & 0.756 & \textbf{0.867}   & 0.815  & 0.833  & 0.789 & 0.810 & 0.955  \\
       &  -0.367   & 0.722  & 0.666  & \textbf{0.888} & 0.761 & 0.756   & 0.879  & 0.861  & 0.842 & 0.900 & 0.953  \\
            & All  & 0.756  & 0.716  & 0.849 & 0.777 & 0.852   & 0.879  & 0.874  & 0.859 & 0.876 & 0.948  \\       
\midrule
ViT S   & All  & \textbf{0.777}  & \textbf{0.761}  & 0.759 & 0.759 & 0.836   & 0.883  & \textbf{0.895}  & 0.873 & 0.882 & \textbf{0.958}  \\
ViT T    & All  & 0.757  & 0.759  & 0.757 & 0.757 & 0.814   & 0.873  & 0.873  & 0.873 & 0.873 & 0.945  \\
\bottomrule
\end{tabular}
\end{table}

\section{EXPERIMENTAL RESULTS}
The experiments evaluated the performance of TST and ViT models under various training subset conditions. We found the highest metrics of the TST model for training on signals with the higher (brightest) flash strength: BA = 0.805 for 1.114 (\(log\; cd.s.m^{-2}\))  flash strength. The second-best accuracy was with ViT trained on wavelets of transformed mixed signals: BA = 0.777. Introducing synthetic signals into the training dataset significantly enhanced the performance of all models. Specifically, the BA of TST trained on the flash strength data at 1.114 (\(log\; cd.s.m^{-2}\)) increased by 10\% and ViT by 13\%.

Following the assessment by clinical experts (D.T. and P.C.), the subsequent deductions were formulated: the generated synthetic ERG waveforms exhibited the key morphological features of the real ERGs recorded in ASD and control subjects. Notably, the high-frequency component of the oscillatory potentials on the ascending limb of the b-wave was present. Additionally, the a- and b-wave amplitudes were in close agreement with the natural ERG signals, although the time to peaks was more variable. 

\begin{figure*}[htp]
  \centering
  \subfigure[Synthetic control signals, 1.114 flash strength.]{\includegraphics[width=10cm]{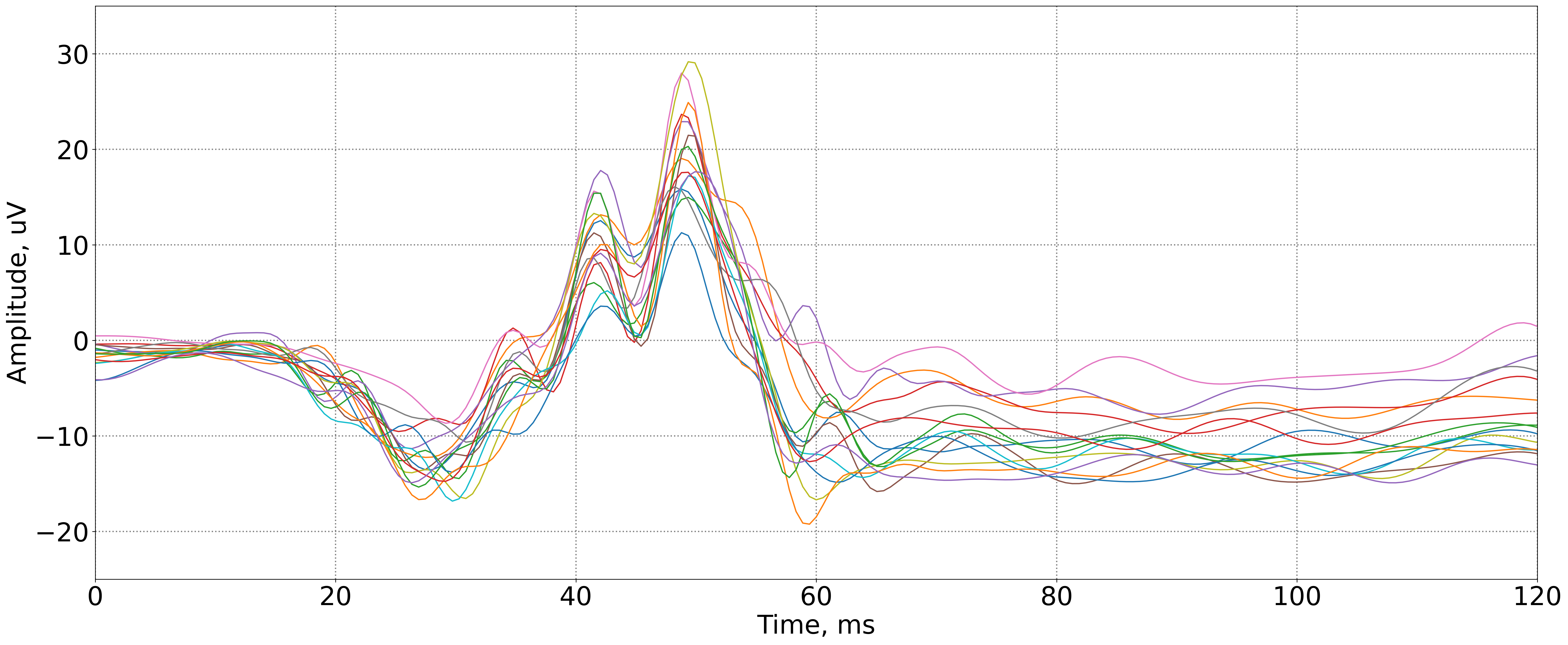}}
  \subfigure[Synthetic control signals, 1.204 flash strength.]{\includegraphics[width=10cm]{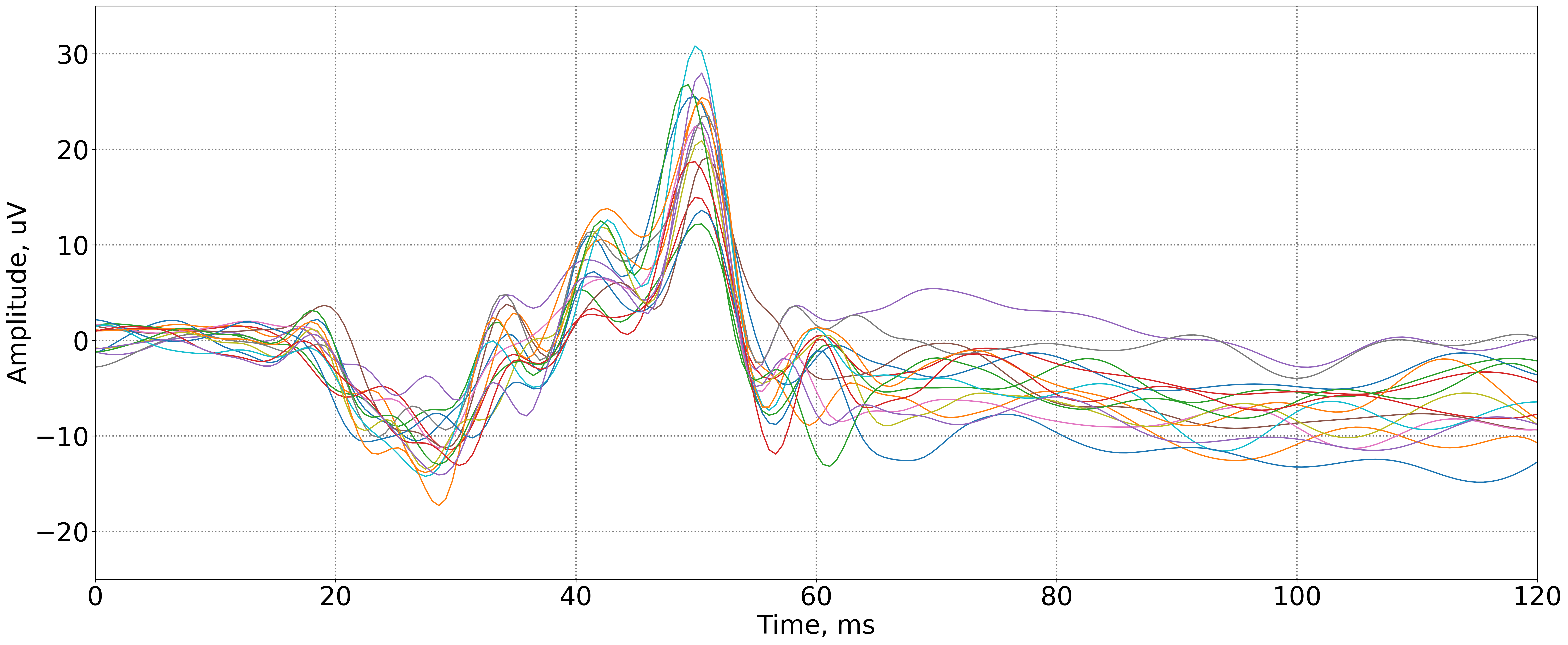}}
  \subfigure[Real control signals, 1.114 flash strength.]{\includegraphics[width=10cm]{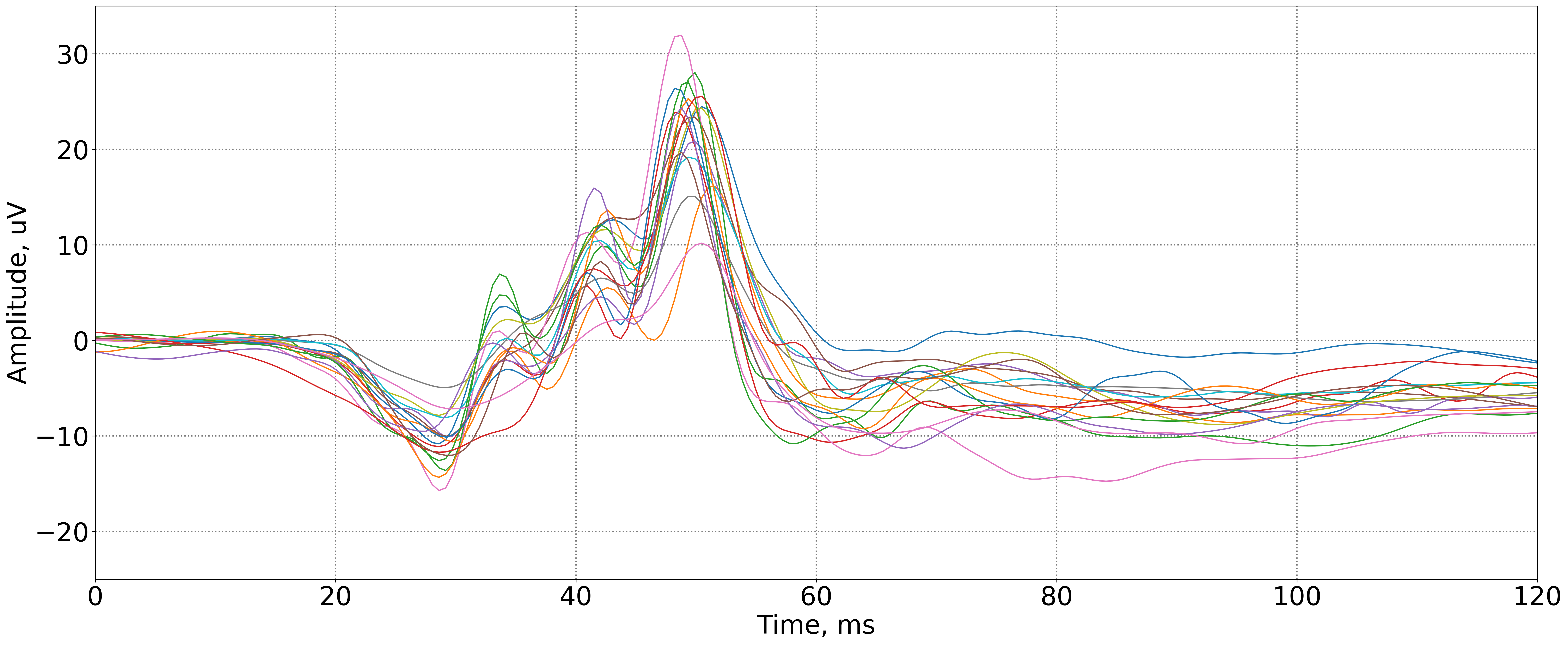}}
  \subfigure[Real control signals, 1.204 flash strength.]{\includegraphics[width=10cm]{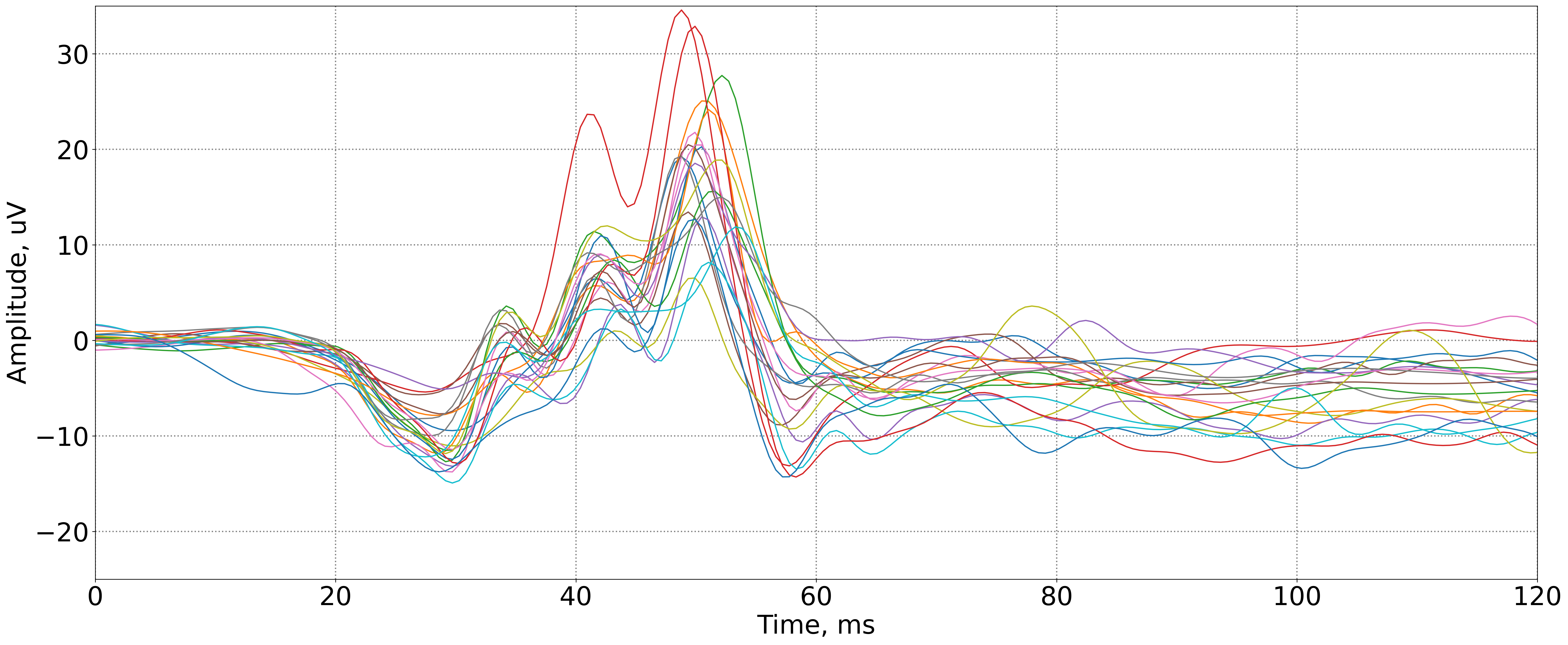}}  
\caption{Series of real and synthetic control ERG signals generated by CGAN and filtered using Butterworth low-pass filter.}
\label{fig:series}
\end{figure*}

Fig. \ref{fig:series} shows examples of Control and ASD series of ERG signals generated using CGAN and then filtered using Butterworth low-pass filter \cite{gauthier2019effects}. Synthetic dataset of 1.204, 1.114, 0.949, and 0.799 (\(log\; cd.s.m^{-2}\)) flash strengths are publicly available on DataPort \cite{npv7-8063-24}.

\section{Conclusions}
Synthetic reference signals can enhance medical operational efficiency by offering a feasible alternative to natural. Synthesizing facilitates dataset expansion within specialized domains, enabling training resource-intensive networks such as transformers. Incorporating synthetic signals generated by the proposed conditional GAN offers a promising solution to address existing challenges in the domain of AI applied to the  ERG. Through expanding the number of samples collected from subjects and generating synthetic waveforms provides a new opportunity to expand AI modeling of the ERG in rare or heterogeneous populations where large datasets are required. The augmentation with synthetic signals will provide the opportunity to train heavy models such as Transformers to support the early detection of retinal disorders. Furthermore, the non-personal nature of synthetic signals permits their open-source publication, making them suitable for sharing without violating patient privacy concerns. We demonstrate the first application of synthetic signals to improve the classification of ASD from controls using signal analysis of the ERG.


%
%
%
\bibliographystyle{splncs04}
\bibliography{refs}
%
\end{document}